\documentclass{article}
\usepackage{ijcai26}

\usepackage{times}
\usepackage{soul}
\usepackage{url}
\usepackage[hidelinks]{hyperref}
\usepackage[utf8]{inputenc}
\usepackage[small]{caption}
\usepackage{graphicx}
\usepackage{amsmath}
\usepackage{amsthm}
\usepackage{booktabs}
\usepackage{algorithm}
\usepackage{algorithmic}
\usepackage{natbib}
\usepackage{subcaption}
\usepackage{amsmath, amssymb, amsthm}

\usepackage{algorithm}
\usepackage{algorithmic}
\usepackage[switch]{lineno}
\usepackage{natbib}
\usepackage{listings}
\usepackage{subcaption}
\usepackage{mathtools}
\usepackage{booktabs}
\usepackage{colortbl}
\usepackage[table]{xcolor}
\usepackage{multirow}

\title{Sim2Act: Robust Simulation-to-Decision Learning via Adversarial Calibration and Group-relative Perturbation}

\author{
Hongyu Cao$^1$
\and
Jinghan Zhang$^2$
\and
Kunpeng Liu$^2$
\and
Dongjie Wang$^3$
\and
Feng Xia$^4$
\and
Haifeng Chen$^5$
\and
Xiaohua Hu$^6$
\and
Yanjie Fu$^1$\\
\affiliations
$^1$Arizona State University\\
$^2$Clemson University\\
$^3$University of Kansas\\
$^4$RMIT University\\
$^5$NEC Laboratories America\\
$^6$Drexel University\\
\emails
hongyuca@asu.edu,
jinghaz@clemson.edu,
kunpenl@clemson.edu,
wangdongjie100@gmail.com,
f.xia@ieee.org,
haifeng@nec-labs.com,
xh29@drexel.edu,
yanjie.fu@asu.edu
}

\begin{document}

\maketitle

\begin{abstract}
In digital twins, simulation-to-decision enables safe decision learning in digital worlds without risking real-world deployments, thus, has become a cornerstone in mission-critical domains (e.g., supply chains and industrial systems).  
However, real-world system data and labels are often noisy, biased, and incomplete, thus, results into unstable decisions when simulators learned from such data are used for policy training.
Existing methods often fall short: 
1) surrogate simulation models tend to be biased particularly in decision-critical regions in which data labels are sparse or biased; 
2) when interacting with biased surrogate simulators, policy learning is sensitive to perturbations. 
It is critical to investigate how to robustify both  simulation and decisions. 
To this end, we propose a robust learning framework to advance
1) \textit{simulation fidelity by adversarial calibration}, 2) \textit{policy robustness with group-relative perturbations}. 
Our solution enables non-disruptive robustness that is stable under perturbation while preserving decision performance. 
We present extensive experiments on multiple supply chain benchmarks (DataCo, GlobalStore, and OAS) to demonstrate the simulation and decision robustness of our method in perturbation settings.
\end{abstract}

\section{Introduction}
In many high-stakes applications such as supply chain management and complex industrial operations,
decision-makers are increasingly trained using learned surrogate simulators
to avoid costly, risky, or privacy-constrained interactions with real environments~\citep{agrawal2025adaptivefewshotlearningafsl}.
This simulation-to-decision paradigm firstly develops a surrogate simulator of the environment: given an action and output next state and reward, then learns optimized policies by interacting with the simulator, among which the Sim2Dec pipeline~\citep{bai2025supplychainoptimizationgenerative}
is a recent instantiation.

Investigating robustness is critical because simulators are often learned from real world noisy or imperfect data . These data imperfections create an issue: while a simulator might look accurate  (e.g., RMSE, MAE) on average in terms of reward prediction, it can exhibit significant biases on rewards over actions; that is, predicted rewards of common actions with large training data are accurate, yet predicted rewards of rare risky actions are less accurate. Such uneven reward predictive accuracies are dangerous for decision-making.  Because policies choose actions by comparing predicted rewards of one action against another,  small reward prediction errors on some actions can flip the ranking orders of all actions, leading to unstable or unsafe decisions.  In this paper, we study the problem of how to robustify both simulation (i.e., surrogate environment modeling) and decision-making (i.e., policy learning). 

Exiting studies of robust decision making are two fold: 1)  from the simulator perspective: how to improve simulation fidelity; 2) from the decision making perspective, how to regularize decision policies under uncertainty.
Despite these advances, two fundamental challenges remain unresolved.
%
On the simulator side, Sim2Dec and other simulation-based methods aim to enhance average predictive accuracy through optimizing simulation fidelity~\citep{YevgenievichBarykin2020}, ensembles~\citep{Correia2023}, physics-informed planning~\citep{bai2025supplychainoptimizationgenerative}, or introducing uncertainty and probabilistic reward~\citep{ATANASSOV20081477}. 
Most simulators reduce average errors in order to be mostly right across all scenarios.
However, in certain state-action pairs, which we call decision-critical regions, small simulation errors can fully distort action ranking orders and significantly reduce policy performance~\citep{fonteneauBatchModeReinforcement2013, NEURIPS2019_2c048d74, zhao2021calibratingpredictionsdecisionsnovel}.
Because the action rewards in simulation are not aligned with true rewards of actions in those decision-critical regions, a tiny $1\%$ error in reward prediction result in much more than a $1\%$ error in decision paths (e.g, a rank reversal).
\textbf{Issue 1 (Simulation-Action Unalignment):}
How can we reduce simulation errors in decision-critical regions while preserving average accuracies, so that small mispredictions can not flip the entire action ranking orders?
On the decision-maker side, 
existing model-based and offline reinforcement learning methods~\citep{huang2022sensitivityanalysisgeneralizationexperimental, liu2021constrainedmodelbasedreinforcementlearning}
typically rely on adversarial perturbations~\citep{pinto2017robustadversarialreinforcementlearning, zhang2021robustdeepreinforcementlearning, zhang2025stateawareperturbationoptimizationrobust, liu2024robustdeepreinforcementlearning}
or conservative regularization schemes~\citep{derman2020distributionalrobustnessregularizationreinforcement, yang2022rorlrobustofflinereinforcement, LI2025126888}
to improve robustness.
However, this often results in ``policy collapse'', where the model becomes too afraid of potential errors that it abandons not just high-risk low-reward, but also high-risk high-reward opportunities entirely. 
Instead of treating every uncertainty as a threat, we need to  distinguish between unacceptable risks and recoverable errors, maintaining robustness without sacrificing the pursuit of high rewards.
\textbf{Issue 2 (Seeing Every Uncertainty as Threat):}  How can we protect a policy against prediction errors without forcing it to become overly timid and discard all high-risk high-reward actions?

\textbf{Our Perspective:}
We propose the Sim2Act framework to address the two issues.
To address Issue~1 (simulation-action unalignment), we introduce adversarial simulator calibration.
Instead of optimizing average simulation accuracy, we identify and re-weight prediction errors of state-action pairs that have high impact on action ranking. Through adversarial reweighting, the simulator is calibrated to prioritize decision-critical regions and align predicted outcomes with real action utility instead of  global prediction metrics~\citep{liu2024robustdeepreinforcementlearning, 10889801}.
To address Issue~2 (seeing every uncertainty as treat), we adopt group-relative decision-making perturbation.
Rather than enforcing pessimistic worst-case constraints, we optimize the policy by preserving relative action preferences across a group. This group-relative formulation stabilizes policy learning under simulator imperfection while maintaining nominal performance~\citep{shao2024deepseekmathpushinglimitsmathematical, zhang2025grpoleaddifficultyawarereinforcementlearning}.
These two technical components  form a robust pipeline to robustify both  simulator surrogate and policy learning.
\noindent\textbf{Our main contributions are as follows:}
\begin{itemize}
    \item We introduce an adversarial simulator calibration method
    that reweighs surrogate outputs based on decision-critical errors, in order to align simulation fidelity with downstream action selection.
    \item We propose a group-relative perturbation strategy
    to stabilize policy preferences and enable robust action selection under simulator uncertainty.
    \item Extensive experiments on the DataCo, GlobalStore, and OAS benchmarks
    show that Sim2Act consistently outperforms existing robustness baselines
    under a range of structured and unstructured perturbations.
\end{itemize}

\begin{figure*}[!t]
    \centering
    \begin{subfigure}[t]{0.28\linewidth}
        \centering
        \includegraphics[height=5cm, width=\linewidth]{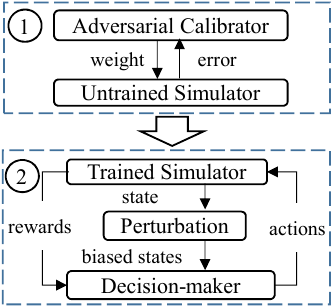}
        \caption{Overview}
        \label{fig:overview}
    \end{subfigure}
    \begin{subfigure}[t]{0.7\linewidth}
        \centering
        \includegraphics[height=5cm, width=\linewidth]{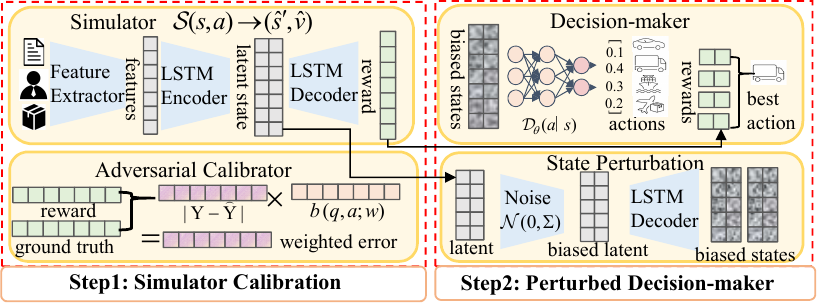}
        \caption{Module Details}
        \label{fig:framework}
    \end{subfigure}
    \caption{\textbf{Sim2Act Framework.} 
    (a) The overview illustrates the two-stage structure of our framework: simulator calibration and perturbed decision-making. (b) Module details: The simulator, implemented as an LSTM-based encoder-decoder model with linear feature extractor, predicts outcomes from state-action pairs and applies adversarial correction to reduce systematic prediction errors. During decision-making, latent states are perturbed using Gaussian noise derived from the simulator's estimated covariance $\Sigma$, producing a distribution of plausible states. The decision-maker $\mathcal{D}_\theta$ learns a robust policy by maximizing group-relative advantages across these perturbed states, enhancing generalization under latent uncertainty.}
    \label{fig:framework_dual}
\end{figure*}

\section{Problem Statement}
\label{sec:problem}

We consider a simulation-based decision-making learning process.
Concretely, the process follows a two-stage pipeline.
First, a simulation model is trained offline from historical data to approximate real-world environment dynamics and rewards.
Second, a decision policy is optimized by interacting with the learned simulator, without access to risky real environment feedback.
Although this approach can achieve high performance under the simulator, policies may behave unreliably at deployment time
when simulator imperfections affect action-sensitive decisions.
Therefore, our objective is not only to maximize expected return under the learned simulator,
but also to ensure robust decision behavior that remains stable under realistic uncertainty.



We now introduce the formal setting and notation.

\textbf{Simulation Model and Decision Policy.}
We model the learned simulator as a parameterized function $\mathcal{S}_{\theta_s}$ trained offline to approximate environment dynamics.
Given a state--action pair $(s_t, a_t)$, the simulator outputs a predicted next state and reward:
$\mathcal{S}_{\theta_s}(s_t, a_t) = (\hat{s}_{t+1}, \hat{r}_t)$.
A decision policy $\mathcal{D}_{\theta_d}$, parameterized by $\theta_d$, maps states to actions,
where $a_t \sim \mathcal{D}_{\theta_d}(\cdot \mid s_t)$.
During policy training, the decision-maker $\mathcal{D}_{\theta_d}$ interacts with trained simulator $\mathcal{S}_{\theta_s}$.

\textbf{Robustness of Decision Policies.}
We characterize robustness as the sensitivity of a policy’s performance to structured perturbations of simulator predictions.
Intuitively, a robust policy should maintain stable behavior under locally constrained variations of simulator outputs.
Formally, we consider perturbed states $s'$
generated via latent-space noise:
$s' \sim s + \mathcal{N}(0, \Sigma)$.
The robustness of a policy $\mathcal{D}_{\theta_d}$
is defined as the expected degradation in predicted rewards as Equation~\eqref{eq:robustness_definition}:
\begin{equation}
\label{eq:robustness_definition}
\text{R}(\mathcal{D}_{\theta_d}) =
- \mathbb{E}_{s}
\left[
\mathcal{S}_{\theta_s}(s, \mathcal{D}_{\theta_d}(s))
-
\mathcal{S}_{\theta_s}(s', \mathcal{D}_{\theta_d}(s'))
\right],
\end{equation}
where $\mathcal{S}_{\theta_s}(\cdot)$ denotes the reward prediction
(i.e., the $\hat{r}$ component of the simulator output).

\textbf{Learning Objective.}
Our goal is to learn a decision policy that both achieves high expected return under the simulator and maintains sufficient robustness.
This leads to the following optimization problem as Equation~\eqref{eq:task_definition}:
\begin{equation}
\label{eq:task_definition}
\theta_d^* =
\arg\max_{\theta_d}
\mathbb{E}
\left[
\sum_{t=0}^{T}
\mathcal{S}_{\theta_s}(s_t, \mathcal{D}_{\theta_d}(s_t))
\right],
\quad
\text{s.t. }
\text{R}(\mathcal{D}_{\theta_d}) > \delta.
\end{equation}

\section{The Sim2Act Approach}

\subsection{An Overview of Sim2Act}

Figure~\ref{fig:framework_dual} shows \textbf{Sim2Act} includes robust simulator learning and robust policy learning: 
\textbf{Step 1: Action-aligned Simulator Calibration} is to reduce simulation errors in decision-critical regions while preserving average accuracies, so that small mispredictions can not flip the entire action ranking orders. 
We introduce an adversarial calibration method to re-weight surrogate outputs based on errors of predicting decision-critical state-action pairs, in order to strengthen the robustness.
\textbf{Step 2: Group-Relative Perturbation} is to  protect a policy against becoming overly timid due to prediction error penalization so that the policy will not see every uncertainty as threat and discard high-performing actions. 
We propose to train the policy  with a group-relative perturbation strategy. In particular, we sample a coherent group of latent perturbations (e.g., from the simulator's learned covariance) around each nominal state. We then compare action rewards across that group, and optimize the policy to preserve relative action rankings within the group.  
This group-wise comparison enables the policy to distinguish and consider high-risk high-reward actions over high-risk low-reward actions, thus, prevent policy from seeing all uncertainty as threat and to maintain robustness.

\subsection{Step 1: Action-Aligned Simulator Calibration}

\noindent\textbf{Why Action-aligned Simulator Calibration Matters?}
In Sim2Act, policy learning relies on a learned simulator to provide state transitions and rewards.  
Compared with traditional simulator training objectives (e.g., global MSE minimization),  the simulator shouldn't just be accurate on average. 
This is because not all mistakes of predicting action rewards are equal. 
The simulator must be most accurate on the specific actions that change the final decision.



\noindent\textbf{The Simulation Calibration Model Structure.}
The simulation calibration structure includes a simulator and a calibrator.

\underline{\textit{1) The simulator.}} 
The simulator (denoted by $\mathcal{S}(s, a) \rightarrow (\hat{s}', \hat{r})$) is a latent-variable model that maps a state-action pair $(s, a)$ to a predicted next state and a reward, given by:
\begin{equation}
\label{eq:simulator}
\mathcal{S}(s, a) = (\hat{s}', \hat{r}) = G(E(s, a)),
\end{equation}
where $E(\cdot)$ encodes the input into a latent representation and $G(\cdot)$ decodes it to environment outcomes.

\underline{\textit{2) The calibrator.}}  We develop a lightweight action-conditioned calibrator (denoted by $\bar{b}(s, a; w)$) that  mimics a policy to output a normalized importance weight for each state-action pair, given by:
\begin{equation}
\label{eq:softmax_attention}
\bar{b}(s, a; w) = \exp(\langle s, w_a \rangle) /
{\sum_{i=1}^{K} \exp(\langle s, w_{a_i} \rangle)}.
\end{equation}
Here, $w_a$ denotes a learnable parameter associated with action $a$. The calibrator modulates simulator training by assigning higher weights to state-action pairs whose prediction errors are more consequential for action ranking, thereby enabling action-aligned calibration in decision-critical regions.

\underline{\textit{3) The interactions between simulator and calibrator.}}
The simulator outputs predicted reward for each action along with corresponding errors (i.e., potential deviation of  predicted reward compared with real reward).  Based on the prediction errors of each action, the calibrator outputs a probability distribution of each action as weights ($\bar{b}(s, a; w)$)), which are used in the gradient descent optimization of simulator to re-weight reward prediction losses.  

\noindent\textbf{The Objective Function of Calibration.}
The simulation calibration can be seen as a mini-max optimization problem, because the calibrator acts as an adversary that identifies the worst-case errors (the max part), forcing the simulator to prioritize fixing the mistakes that matter most for decision-making (the mini part).
In particular, in the max part, the calibrator scans the data to find where the simulator is most wrong about action outcomes, then shine a spotlight on those errors by giving them the highest weights; in the mini part, the simulator then tries to minimize the errors that the calibrator has highlighted, ensuring it becomes accurate in the most critical areas rather than being average accurate. This minimax objective is given by: 
\begin{equation}
\label{eq:calibration_objective}
J(w, \theta_s) = \sum_{i=1}^K \mathbb{E}_{(x,y)\in D} \left[ \bar{b}(s, a_i; w) \cdot \left\| y - \mathcal{S}(x; \theta_s) \right\| \right],
\end{equation}
where $x = (s, a)$, $y = (s', r)$ are ground-truth transitions from the offline dataset $D$. 
Here, the calibrator $w$ seeks to maximize this objective (to increase the weights of high-impact errors for actions), while the simulator $\theta_s$ seeks to \textit{minimize} it (to correct those errors). 

\noindent\textbf{Solving the Optimization Problem.}
We employ an alternating optimization strategy to stabilize simulator training. 
This process iteratively refines the focus of the calibrator and the accuracy of the simulator:
First, we freeze the simulator parameters $\theta_s$
and update the calibrator $w$ via gradient ascent
during the optimization process
as Equation~\eqref{eq:caliberator_update}.
\begin{equation}
\label{eq:caliberator_update}
w^{(t)} \leftarrow w^{(t-1)} + \eta_w \nabla_w J(w, \theta_s^{(t-1)}).
\end{equation}
This step identifies high-error regions in the current simulator by increasing the weights $\bar{b}(s, a; w)$ for samples with large residuals.
We next freeze $w$ and update the simulator parameters $\theta_s$ via gradient descent as Equation~\eqref{eq:update_rule}:
\begin{equation}
\label{eq:update_rule}
\theta_s^{(t)} \leftarrow \theta_s^{(t-1)} - \eta_s \nabla_{\theta_s} J(w^{(t)}, \theta_s).
\end{equation}
By minimizing the weighted loss, the simulator is forced to reduce errors in the regions highlighted by the calibrator.
This alternating procedure continues until the validation error stabilizes, effectively balancing global accuracy with local fidelity in critical regions.

\noindent\textbf{Theoretical Insight for Step 1}
Action-aligned simulator calibration prioritizes decision-critical regions,
tightening the deployment-time performance gap
by aligning simulated rewards with action ranking for decision making.
Formal statements and proof are in Appendix B.

\subsection{Step 2: Group-relative Perturbation}

\noindent\textbf{Why Group-Relative Perturbation Matters?}
%
Perturbations matter because a learned simulator is different messy real world, and perturbations expose whether a policy is dangerously over-optimized to a simulator’s specific imperfections. 
Group-relative perturbations matter because instead of reacting to single noisy perturbed state, they train the policy to compare actions across a local group of perturbations. This can stabilize relative action preferences, reduce overreaction, and keep the policy robust without becoming overly conservative.
By maximizing the compared rewards, the policy effectively learns to task risky action when reward is high.

\noindent\textbf{The  Model Structure.}

\ul{\textit{1) The decision-maker}}
The decision-maker $\mathcal{D}_{\theta_d}$
is a stochastic policy that maps a state
to a distribution over actions:
$\mathcal{D}_{\theta_d}(a \mid s)$.
The policy outputs action distribution and get predictive rewards
from calibrated simulator for policy learning
and then update policy.

\ul{\textit{2) Perturbed states and perturbation groups.}}
We generate perturbations in the simulator's latent space.
Specifically, given a state-action pair $(s, a)$, the simulator predicts a latent distribution characterized by a centroid $z$ and a covariance matrix $\Sigma(s, a)$.
We define a perturbation group as a set of $M$ latent vectors sampled from this distribution:
$\{z_i\}_{i=1}^M$, where $z_i \sim \mathcal{N}(z, \Sigma)$.
These latent vectors are then decoded back into a set of perturbed states $\tilde{S} = \{\tilde{s}_1, \dots, \tilde{s}_M\}$.

\ul{\textit{3) The interaction between decision-maker and perturbation groups.}}
For every decision step, the policy is exposed to the perturbed states of a perturbation group, and generate appropriate action for each state.
The policy then select the best action with highest reward from the actions and update its parameters with selected state-action pairs.

\noindent\textbf{The Objective Function of Perturbation.}
We design a composite loss function to balance robust exploration with utility maximization. 
It consists of a group-relative advantage term and a utility alignment term.
We denote the reward prediction component of the simulator as $\mathcal{S}^r(s, a) \in \mathbb{R}$.
To encourage the policy to favor actions that are robustly superior, we formulate a Group-Relative Advantage.
Here, the performance of a specific action $a_i$ is evaluated not in absolute terms, but relative to the group average $\bar{r}$ of the perturbed neighborhood, which serves as a local baseline:
$\bar{r} = \frac{1}{M} \sum_j \mathcal{S}^r(\tilde{s}_j, a_j)$.
The robust advantage loss is defined as Equation~\eqref{eq:group_adv}:
\begin{equation}
\label{eq:group_adv}
\mathcal{L}_{\text{group-adv}} = - \frac{1}{M} \sum_{i=1}^{M} \underbrace{(\mathcal{S}^r(\tilde{s}_i, a_i) - \bar{r})}_{\text{Group Advantage}} \cdot \log \mathcal{D}_{\theta_d}(a_i \mid \tilde{s}_i).
\end{equation}
It encourages the policy to increase the probability of actions that perform better than the local average under uncertainty,
while suppressing actions that fall below the group mean.
Simultaneously, we define a target reference $r^*$ (e.g., an expert baseline or a maximization target) and minimize the ``regret'' gap $(r^* - \mathcal{S}^r(s, a))$.
This forces the policy to not just be stable, but to push the absolute predicted reward $\mathcal{S}^r(s, a)$ towards the ideal outcome.
The final objective combines these two goals as Equation~\eqref{eq:composite_loss}:
\begin{equation}
\label{eq:composite_loss}
\mathcal{L}_{\text{decision}} = \eta \cdot \mathcal{L}_{\text{group-adv}} + \underbrace{(r^* - \mathcal{S}^r(s, a \sim \mathcal{D}_{\theta_d}))}_{\text{Utility Gap}},
\end{equation}
where $\eta > 0$ balances the stability induced by the group advantage with the aggressiveness of utility maximization.

\noindent\textbf{Solving The Optimization Problem.}
The training process alternates between perturbed states generation and policy update.
In each iteration, we first map a batch of states $s$ to the simulator's latent space to capture local uncertainty (covariance $\Sigma$).
By sampling $M$ latent perturbations, we generate a \textit{group of perturbed contexts} $\tilde{s}_i$ around the nominal state.
Next, the policy acts on this entire group.
We calculate the group mean $\bar{r}$ to determine the \textit{relative advantage} of each action ($\mathcal{L}_{\text{group-adv}}$) while simultaneously monitoring the absolute utility gap on the original state.
Finally, $\theta_d$ is updated via gradient descent to minimize the joint objective, ensuring the policy is both robust to variance and high-performing on average.
The policy parameters $\theta_d$ are updated via gradient descent to minimize the joint loss $\mathcal{L}_{\text{decision}}$.
This procedure repeats until the policy converges to a solution that is both high-performing and robust to the modeled variance. Detailed algorithmic steps are provided in Appendix F.

\noindent\textbf{Theoretical Insights for Step 2.}
Group-relative learning acts as a control variate,
reducing optimization variance while preventing over-conservative policy collapse.
Formal statements and proofs are in Appendix B.



\begin{table*}[htbp]
\centering
\setlength{\tabcolsep}{4pt}
\begin{tabular}{lccccccccccccc}
\toprule
\textbf{Dataset}
& \multicolumn{4}{c}{\textbf{DataCo (43, 165445)}} 
& \multicolumn{4}{c}{\textbf{GlobalStore (27, 51290)}} 
& \multicolumn{4}{c}{\textbf{OAS (22, 28136)}} \\
\cmidrule{1-1} \cmidrule(lr){2-5} \cmidrule(lr){6-9} \cmidrule(lr){10-13}
\textbf{Sim Acc} 
& Risk & Time & Status & Overall 
& Risk & Time & Status & Overall 
& Risk & Time & Status & Overall \\
\midrule
Markov     & 0.4978 & 0.1487 & 0.5040 & 0.3835 
      & 0.4961 & 0.1355 & 0.4934 & 0.3750 
      & 0.5100 & 0.0011 & 0.5068 & 0.3393 \\
Prediction     & 0.7019 & 0.3395 & 0.8161 & 0.6191 
      & 0.8440 & 0.6767 & 0.8430 & 0.7879 
      & 0.7157 & 0.3706 & 0.7510 & 0.6124 \\
Generation     & 0.7024 & 0.3485 & 0.8156 & 0.6221 
      & 0.9366 & 0.8066 & 0.9355 & 0.8929 
      & 0.7149 & 0.3916 & 0.7503 & 0.6189 \\
S2D   & \underline{0.9508} & \underline{0.8851} & \textbf{0.9695} & \underline{0.9351} 
      & \textbf{0.9743} & \underline{0.9255} & \textbf{0.9756} & \underline{0.9585} 
      & \underline{0.7215} & \textbf{0.3985} & \underline{0.7574} & \underline{0.6258} \\
S2A(Ours)  & \textbf{0.9563} & \textbf{0.8875} & \underline{0.9618} & \textbf{0.9352} 
      & \underline{0.9723} & \textbf{0.9744} & \underline{0.9750} & \textbf{0.9650} 
      & \textbf{0.7270} & \underline{0.3937} & \textbf{0.7629} & \textbf{0.6279} \\
\addlinespace[2pt]
\arrayrulecolor{black}\specialrule{1.2pt}{0pt}{0pt}
\addlinespace[2pt]
\textbf{Dec Rwd} 
& T$^{\text{Timely}}$ & T$^{\text{Profit}}$ & Diff & Overall 
& T$^{\text{Timely}}$ & T$^{\text{Profit}}$ & Diff & Overall 
& T$^{\text{Timely}}$ & T$^{\text{Profit}}$ & Diff & Overall \\
\cmidrule{1-1} \cmidrule(lr){2-5} \cmidrule(lr){6-9} \cmidrule(lr){10-13}
Real   & 0.5244 & 0.0364 & 0.4880 & 0.5608 
       & 0.3320 & 0.0848 & 0.2472 & 0.4168 
       & 0.4800 & 0.0000 & 0.4800 & 0.4800 \\
LP     & 0.5162 & 0.5434 & \underline{0.0272} & 1.0596 
       & 0.3552 & 0.6001 & \textbf{0.2449} & 0.9554 
       & 0.5037 & 0.1043 & 0.3994 & 0.6080 \\
DQN    & 0.5276 & 0.2071 & 0.3205 & 0.7347 
       & 0.2827 & 0.9326 & 0.6499 & 1.2153 
       & 0.4817 & 0.0000 & 0.4817 & 0.4817 \\
PPO    & 0.5343 & 0.0000 & 0.5343 & 0.5343 & 0.3476 & 0.0004 & 0.3472 & 0.3480 & 0.4865 & 0.0000 & 0.4865 & 0.4865 \\
GPT3.5    & 0.5258 & 0.2459 & 0.2800 & 0.7717 
       & 0.3298 & 0.0439 & \underline{0.2859} & 0.3736 & 0.4844 & 0.0000 & 0.4844 & 0.4844 \\
S2D    & \underline{0.5397} & \underline{0.5637} & \textbf{0.0240} & \underline{1.1034} & \underline{0.3446} & \underline{0.9278} & 0.5828 & \underline{1.2724} & \textbf{0.4882} & \underline{0.1611} & \underline{0.3271} & \underline{0.6493} \\
S2A(Ours)   & \textbf{0.5447} & \textbf{0.5786} & 0.0339 & \textbf{1.1232} & \textbf{0.3446} & \textbf{0.9460} & 0.6014 & \textbf{1.2906} & \underline{0.4830} & \textbf{0.1886} & \textbf{0.2944} & \textbf{0.6717} \\
\bottomrule
\end{tabular}
\caption{Simulation and Decision Performance Comparison (Refer to Appendix J
\label{tab:simulation_decision_performance}
for standard deviations).}
\end{table*}
\section{Experimental Results}

We conduct extensive experiments on various datasets to evaluate the performance of our method. 
Specifically, our experiments aim to answer: 
Q1: Can our method outperform baselines on the robustness under random or latent perturbations? 
Q2: Aside from robustness, can our method achieve comparable accuracies of in simulation and decision-making like other strong baselines?
Q3: How can we understand and interpret the mechanism and influence of our algorithm?

\subsection{Experimental Setup}

\paragraph{Datasets.} 
We evaluate our approach on three open-source supply chain datasets:  1) \textbf{DataCo}~\citep{ConstanteEtAl2019}, 2) \textbf{Global-Store}~\citep{dataset-global-superstore}, and 3) \textbf{OAS}~\citep{OAS} that cover diverse logistics and shipping scenarios with order-level records, including products, shipment modes, and delivery outcomes.  
We use supply chains as experiment applications because they are a representative industrial system for simulation and decision science studies. 
Each dataset is split into training, validation, test sets (8:1:1), using the same protocol as Sim2Dec~\citep{bai2025supplychainoptimizationgenerative} to ensure a fair comparison. 
To prevent overfitting, we use early stopping and $\ell_2$ weight penalties.  
Both simulator and decision models are trained on the training set and evaluated on the test set.  
Dataset statistics are reported in the sub-captions of Table~\ref{tab:simulation_decision_performance}
in the format \emph{(number of features, number of items)}.

\paragraph{Evaluation Metrics.}
The performance of a simulator is evaluated based on three tasks of predicting delay risk, delivery time, and on-time status, with test-set accuracy averaged across these outputs.
To assess simulator robustness, we use: 
(i) worst-case accuracy, defined by the minimum Overall score across runs with the same perturbation level; 
(ii) variance, defined by the standard deviation of overall scores at the same perturbation level; and 
(iii) drop rate, defined by the average decline from the unperturbed setting.
The performance of a decision-maker is measured by the two metrics: average profit and on-time rate (both normalized to $[0, 1]$). 
We report the (i) absolute difference (denoted by Diff) to capture imbalance, and (ii)  sum (denoted by Overall) of the average profit and on-time rate as total reward, ranging in $[0, 2]$.
To assess decision robustness evaluation, we examine how the Overall metric drops as the perturbation magnitude increases relative to the unperturbed baseline at $\epsilon=0$, rather than relying on single-point measurements.
To evaluate safety under the worst-case scenarios, we report \textbf{Conditional Value at Risk (CVaR@5\%)}, which measures the expected return of the worst 5\% of trajectories. 
All robustness experiments are conducted under two controlled settings: \textbf{Nominal ($p=0$)} and \textbf{Perturbed ($p=0.5$)}, In the Perturbed setting, we introduce latent Gaussian noise to represent severe, structured distribution shifts, while the perturbation sweep probes increasingly adverse conditions.

\paragraph{Baseline Algorithms.} 
We consider three paradigms for \textbf{simulation}:
1) \noindent\textbf{Markov-based simulation}~\citep{gagniuc2017markov}, which models transitions using predefined probabilities.
2) \noindent\textbf{Prediction-based simulation}~\citep{Caruana1997}, which uses multi-task learning to separately predict status variables.
3) \noindent\textbf{Generation-based simulation}~\citep{gu2018nonautoregressiveneuralmachinetranslation}, a non-autoregressive model that jointly generates multiple order features in a single forward pass.

For \textbf{decision-making}, we compare our method against standard and robust optimization strategies:
1) \noindent\textbf{Linear Programming (LP)}~\citep{dantzig2002linear}, a rule-based optimization approach serving as a deterministic baseline.
2) \noindent\textbf{Deep Q-Network (DQN)}~\citep{Mnih2015} and \textbf{PPO}~\citep{schulman2017proximalpolicyoptimizationalgorithms}, representing standard value-based and policy-gradient reinforcement learning methods, respectively.
3) \noindent\textbf{ChatGPT-3.5}~\citep{brown2020languagemodelsfewshotlearners}, evaluated under a zero-shot setting to assess large language model capabilities in supply chain logic.
4) \noindent\textbf{RARL}~\citep{pinto2017robustadversarialreinforcementlearning}.
5) \textbf{EPOpt}~\citep{rajeswaran2017epoptlearningrobustneural}, two representative robust RL methods that utilize adversarial training and ensemble-based generalization, respectively, serving as key benchmarks for robustness evaluation.
6) \noindent\textbf{S2D} (Sim2Dec)~\citep{bai2025supplychainoptimizationgenerative}, the state-of-the-art simulation-to-decision framework and the backbone of our method, which we aim to improve upon.
7) \noindent\textbf{S2D-l}: a variant of S2D, perturbed under latent-structured perturbation), and 
8) \noindent\textbf{Ours-l}: a variant of Sim2Act, perturbed under latent-structured perturbation.
Implementation details, selection rationale, Hyper-parameters and environment are detailed in Appendix H.


\subsection{Experimental Results}

\noindent\textbf{Q1: A Study of Simulation and Decision Robustness Under Perturbations.}

\begin{figure*}[htbp]
    \centering
    \begin{subfigure}{0.32\linewidth}
        \centering
        \includegraphics[width=\linewidth]{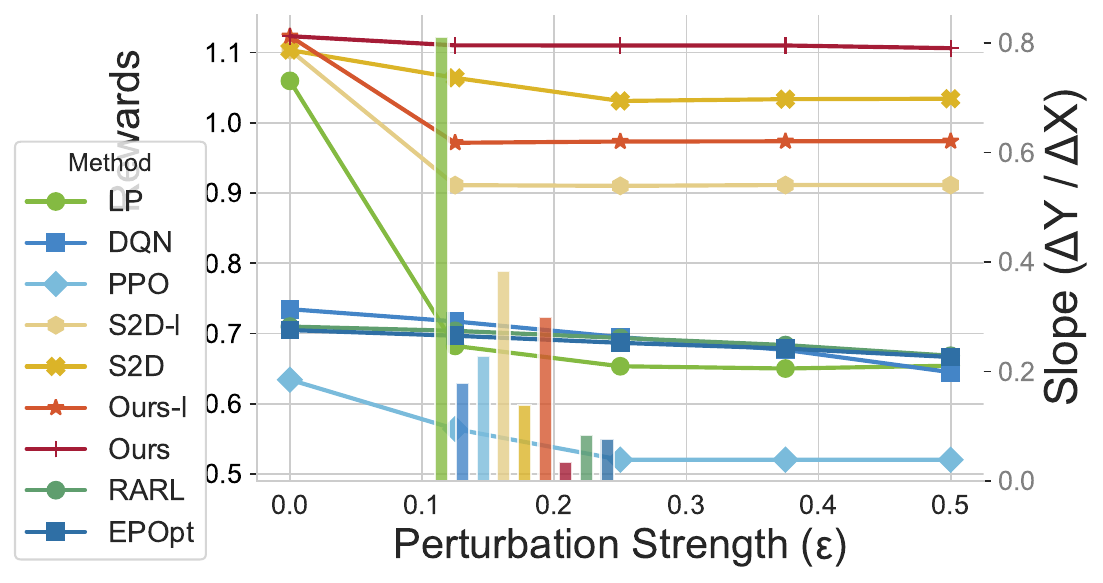}
        \caption{Decision-maker: DataCo (43, 165445)}
        \label{fig:dm_dataco_robustness}
    \end{subfigure}
    \begin{subfigure}{0.32\linewidth}
        \centering
        \includegraphics[width=\linewidth]{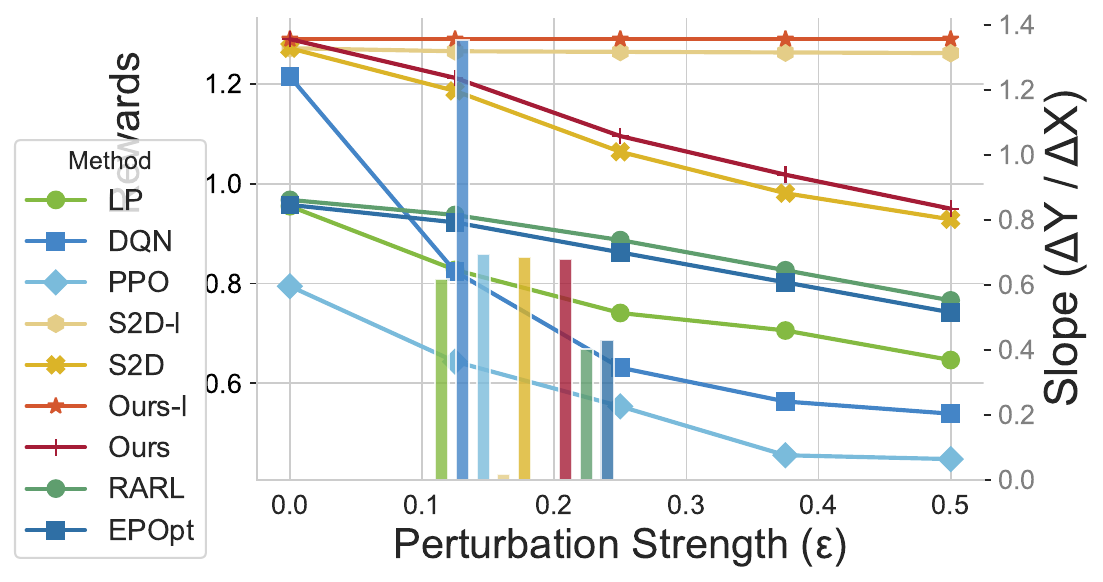}
        \caption{Decision-maker: GlobalStore}
        \label{fig:dm_globalstore_robustness}
    \end{subfigure}
    \begin{subfigure}{0.32\linewidth}
        \centering
        \includegraphics[width=\linewidth]{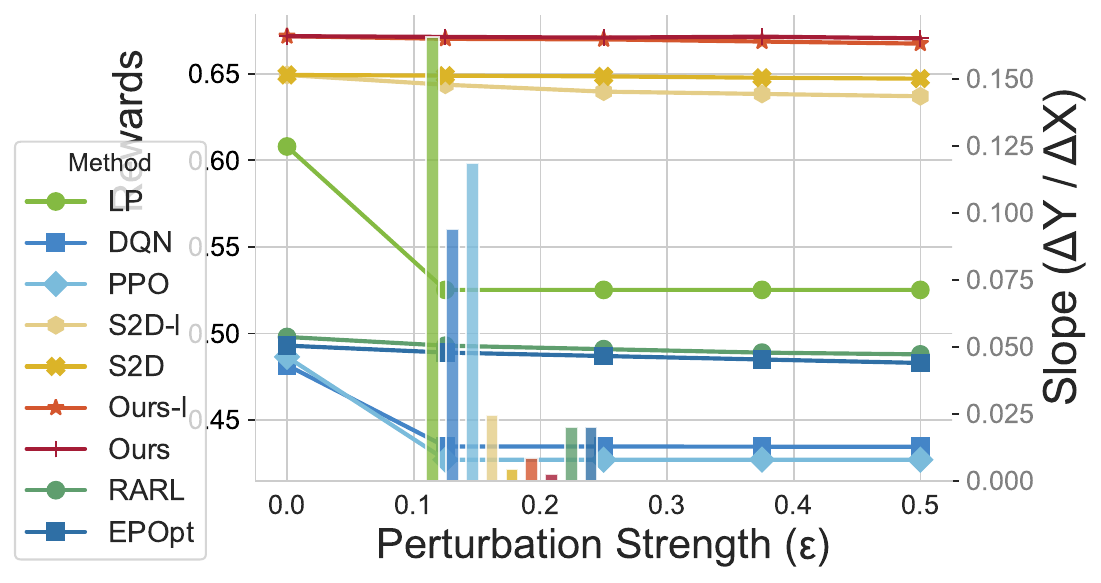}
        \caption{Decision-maker: OAS}
        \label{fig:dm_oas_robustness}
    \end{subfigure}
    \caption{
    Decision-maker robustness under increasing perturbation across three datasets.
    Curves show the degradation of overall decision reward as perturbation strength increases,
    with slopes indicating sensitivity to uncertainty.
    Sim2Act maintains flatter degradation curves and smaller slopes than baselines,
    demonstrating stable performance under both latent-structured and unstructured perturbations
    (Goal~2).
    }

    \label{fig:dm_robustness}
\end{figure*}

\begin{figure}[t] 
    \centering
    \includegraphics[width=\linewidth]{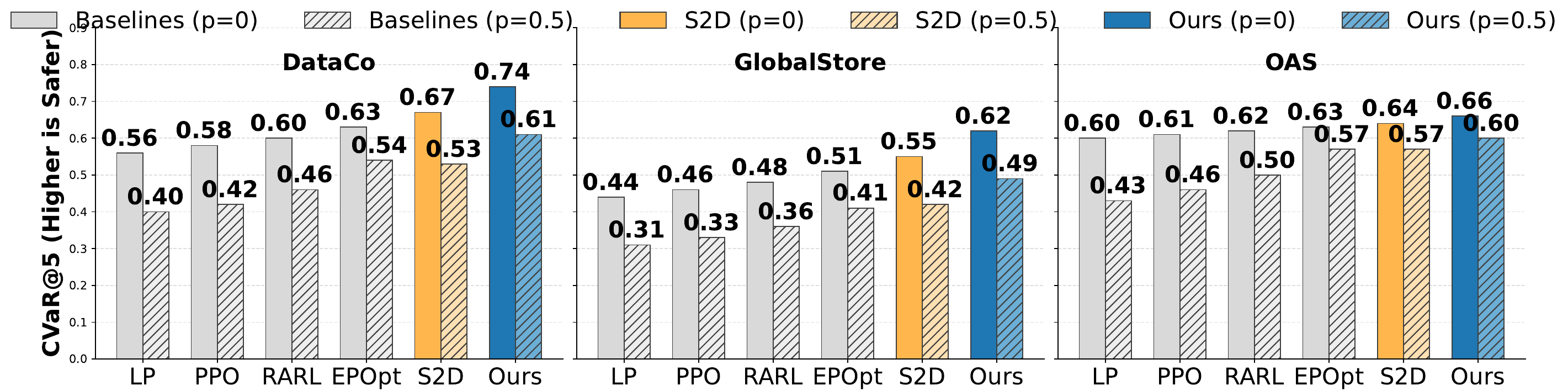}
    \caption{CVaR@5 Robustness on DataCo, GlobalStore, and OAS datasets. 
    Solid bars denote nominal performance ($p=0$), while hatched bars denote performance under perturbation ($p=0.5$). }
    \label{fig:cvar_robustness}
\end{figure}

\begin{figure}[t]
    \centering
    \begin{subfigure}{0.32\linewidth}
        \centering
        \includegraphics[width=\linewidth]{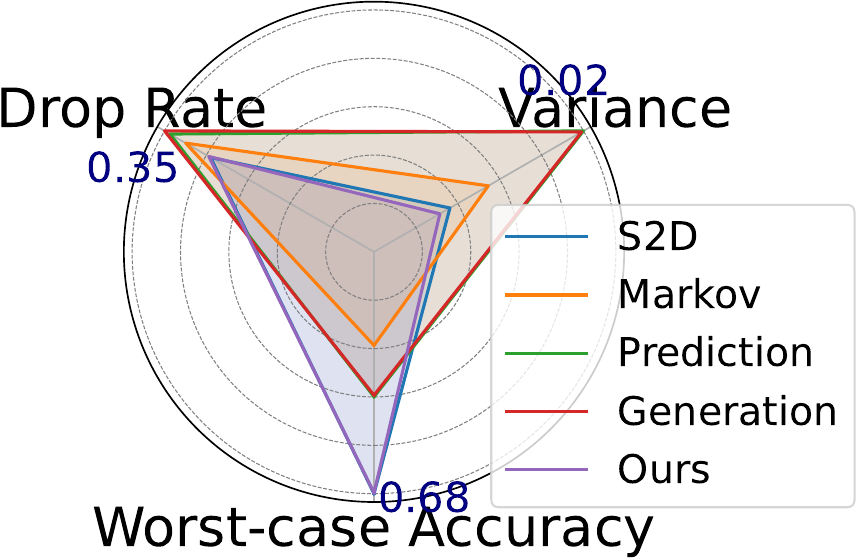}
        \caption{Sim: DataCo}
        \label{fig:sim_dataco_robustness}
    \end{subfigure}
    \begin{subfigure}{0.32\linewidth}
        \centering
        \includegraphics[width=\linewidth]{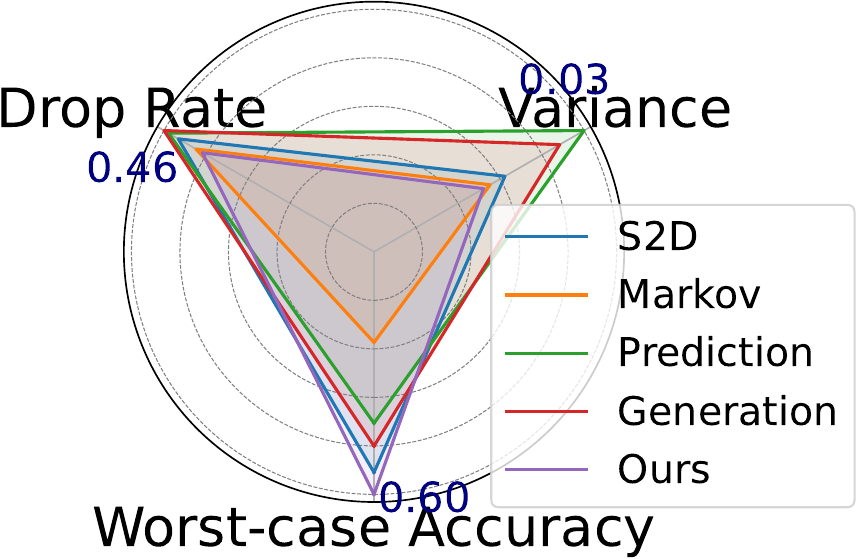}
        \caption{Sim: GlobalStore}
        \label{fig:sim_globalstore_robustness}
    \end{subfigure}
    \begin{subfigure}{0.32\linewidth}
        \centering
        \includegraphics[width=\linewidth]{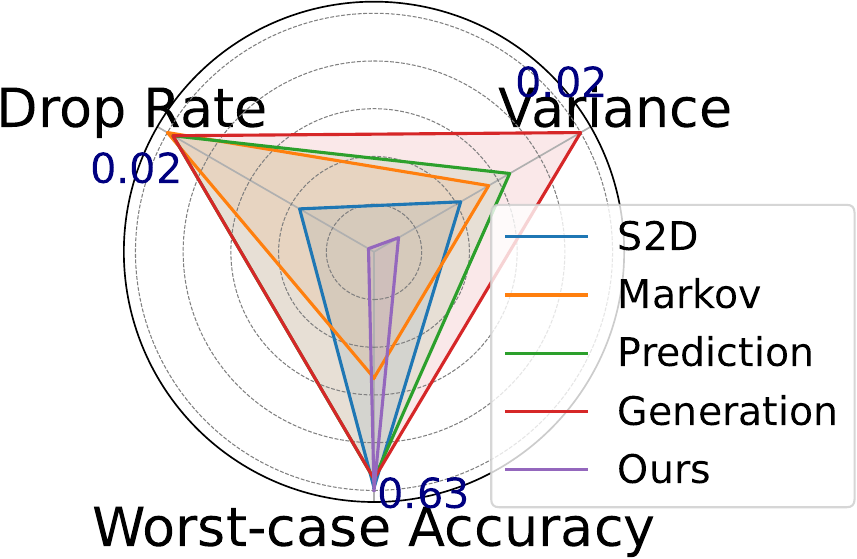}
        \caption{Sim: OAS}
        \label{fig:sim_oas_robustness}
    \end{subfigure}  
    \caption{
    Simulator robustness under perturbation across three datasets.
    Radar plots summarize worst-case accuracy, variance, and drop rate,
    Sim2Act improves worst-case accuracy and reduces variability,
    demonstrating effective decision-critical calibration.
    }
    \label{fig:sim_robustness}
\end{figure}

\noindent\underline{A: Decision Robustness.}
To assess decision robustness, we conduct sensitivity tests using two complementary perturbation mechanisms that correspond to realistic deployment failure modes: \textbf{latent-structured perturbation} (denoted by the \texttt{-l} suffix), which samples Gaussian perturbations in the simulator latent space guided by the learned covariance, and \textbf{random input perturbation} (denoted by no suffix), which applies unstructured noise directly to inputs. Implementation details remain in Appendix K.
%
These two mechanisms serve as representative \emph{probes} of failure behavior (structured vs.\ unstructured) and are evaluated across a range of magnitudes so that we can characterize whether decision performance drop is gradual (graceful) or abrupt (catastrophic). 
We observe that Sim2Act does not rely on any specific noise distribution: calibration and group-relative mechanisms implemented by correcting decision-critical inconsistencies and stabilizing relative action preferences, rather than by tailoring to a single perturbation family.
\noindent\textbf{Average Performance Stability.} Figure~\ref{fig:dm_robustness} highlights that our method maintains stable performance under both latent-structured and random perturbations. 
On DataCo, our approach shows almost no degradation ($1.1232 \rightarrow 1.1222$) while S2D drops notably ($1.1034 \rightarrow 1.0342$). 
On GlobalStore, \texttt{Ours-l} remains invariant around $1.29$, whereas baselines degrade substantially. On OAS, our method preserves near-constant rewards ($0.6717 \rightarrow 0.6705$), outperforming both S2D and LP. 

\noindent\textbf{Risk Analysis (CVaR).} Beyond average returns, we explicitly evaluate policy safety using CVaR@5(Figure~\ref{fig:cvar_robustness}). While S2D shows competitive nominal performance ($p=0$), it becomes brittle under perturbation, suffering a 20.9\% drop on DataCo ($0.67 \to 0.53$) at $p=0.5$. In contrast, Sim2Act demonstrates robust risk control, maintaining a high CVaR of 0.61, which validates that our group-relative mechanism effectively mitigates tail risks caused by simulation errors.
These results confirm that Sim2Act achieves robust and non-disruptive performance across structured and unstructured perturbations. Full detailed degradation curves and distributional analyses, are in Appendix D.

\noindent\underline{B: Simulation Robustness.}
Figure~\ref{fig:sim_robustness} shows that our method achieves consistently higher \emph{worst-case accuracy} across all datasets, e.g., $0.679$ on DataCo compared to $0.225$ for Markov. 
Besides, it yields the lowest performance variance that highlights that calibration improves simulator reliability in decision-critical regions. Complete variance and drop-rate statistics are reported in Appendix J.


\noindent\textbf{Q2: Simulation and Decision Performance Comparison.}

\noindent\underline{A: Simulation Accuracy.}
As illustrated in Figure~\ref{tab:simulation_decision_performance}, we observe that S2A achieves not just comparable simulation accuracy to S2D but also improve downstream decision quality. 
On DataCo, the profit score improves from $0.5637$ to \textbf{0.5786}, and the overall decision score increases from $1.1034$ to \textbf{1.1232}. 
Similar trends are observed across other datasets, highlighting that localized surrogate corrections enhance fidelity in decision-critical regions. Detailed per-metric results are provided in Appendix J.

\noindent\underline{B: Decision Reward.}
As illustrated in Figure~\ref{tab:simulation_decision_performance}, our method not just enhances decision robustness but also preserve decision reward. On \textbf{GlobalStore}, S2A boosts profit from $0.9278$ to \textbf{0.9460} while maintaining timeliness at $0.3446$. 
On \textbf{OAS}, the overall decision score rises from $0.6493$ to \textbf{0.6717}, indicating improved robustness under perturbations. 
These results demonstrate that S2A fulfills our design goals by improving decision-critical calibration and enabling robust yet high-quality policy behavior in offline decision settings.

\begin{figure*}[t]
    \centering
    \begin{subfigure}{0.25\linewidth}
        \centering
        \includegraphics[width=\linewidth]{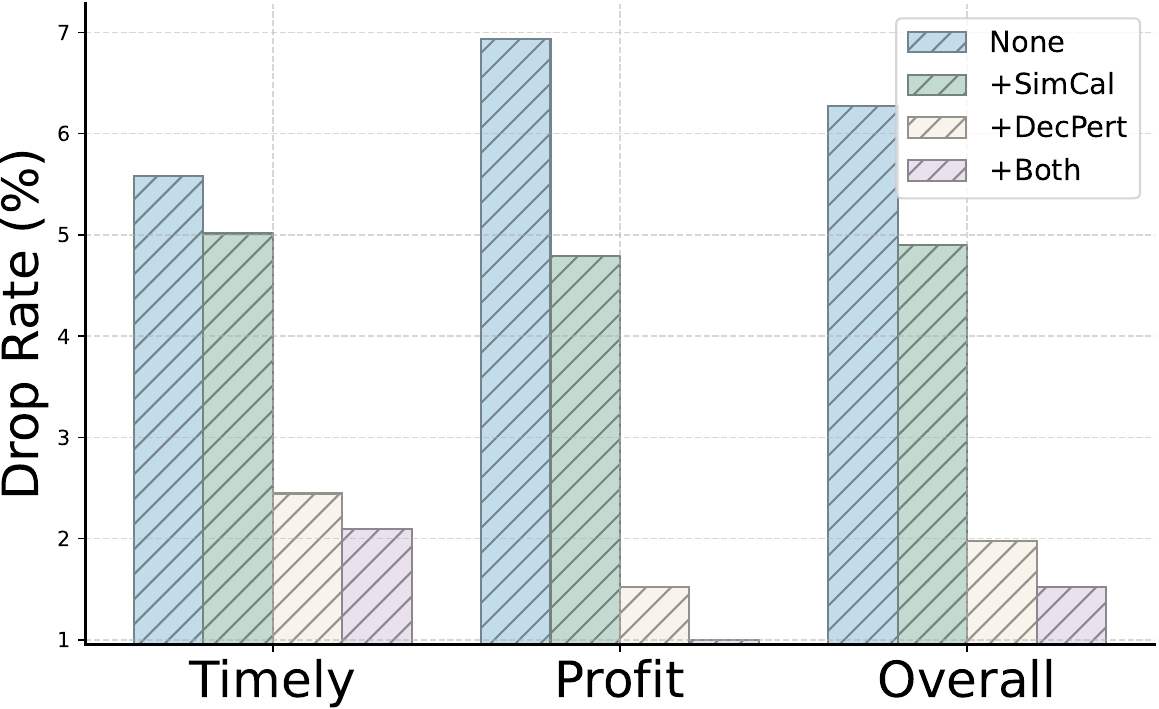}
        \caption{DataCo Ablation}
        \label{fig:DataCo_dm_robustness_ablation_study}
    \end{subfigure}\hfill
    \begin{subfigure}{0.25\linewidth}
        \centering
        \includegraphics[width=\linewidth]{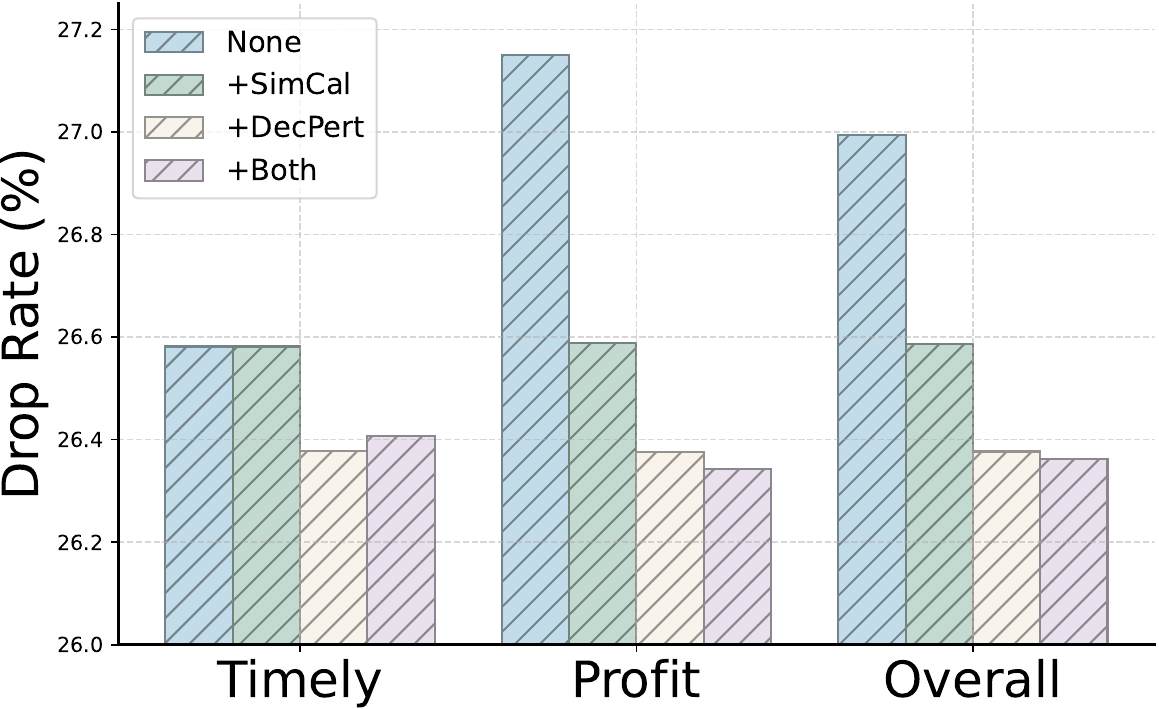}
        \caption{GlobalStore Ablation}
        \label{fig:GlobalStore_dm_robustness_ablation_study}
    \end{subfigure}\hfill
    \begin{subfigure}{0.25\linewidth}
        \centering
        \includegraphics[width=\linewidth]{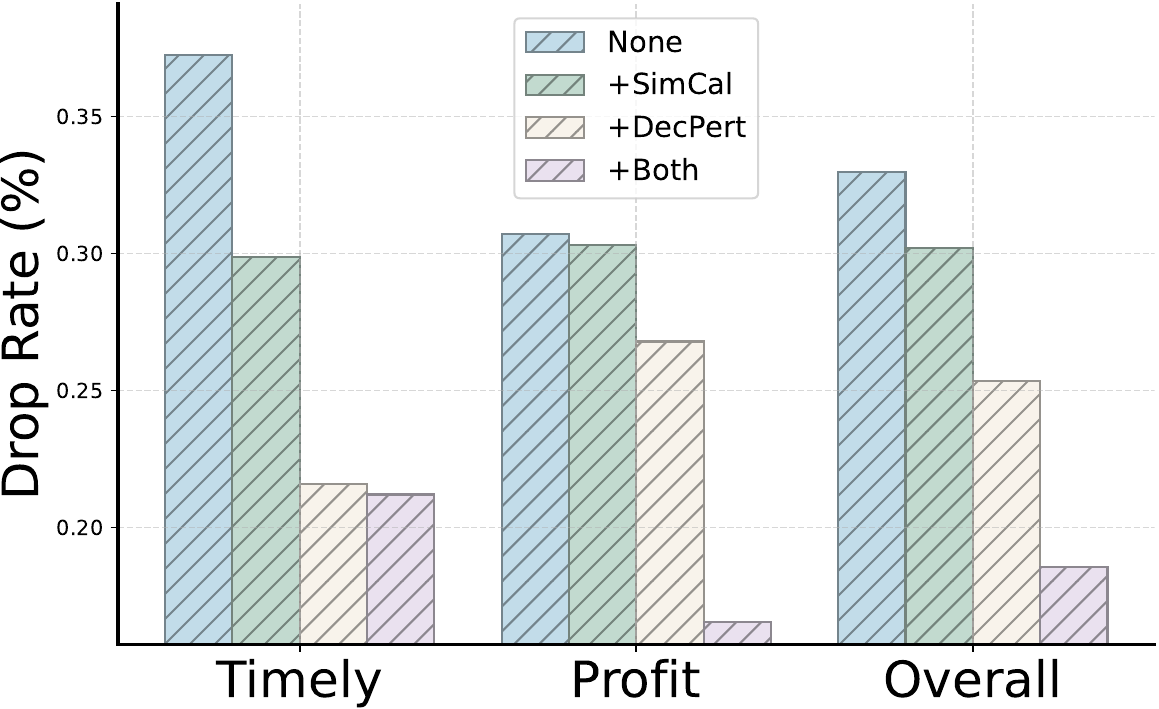}
        \caption{OAS Ablation}
        \label{fig:OAS_dm_robustness_ablation_study}
    \end{subfigure}\hfill
    \begin{subfigure}{0.2\linewidth}
        \centering
        \includegraphics[width=\linewidth]{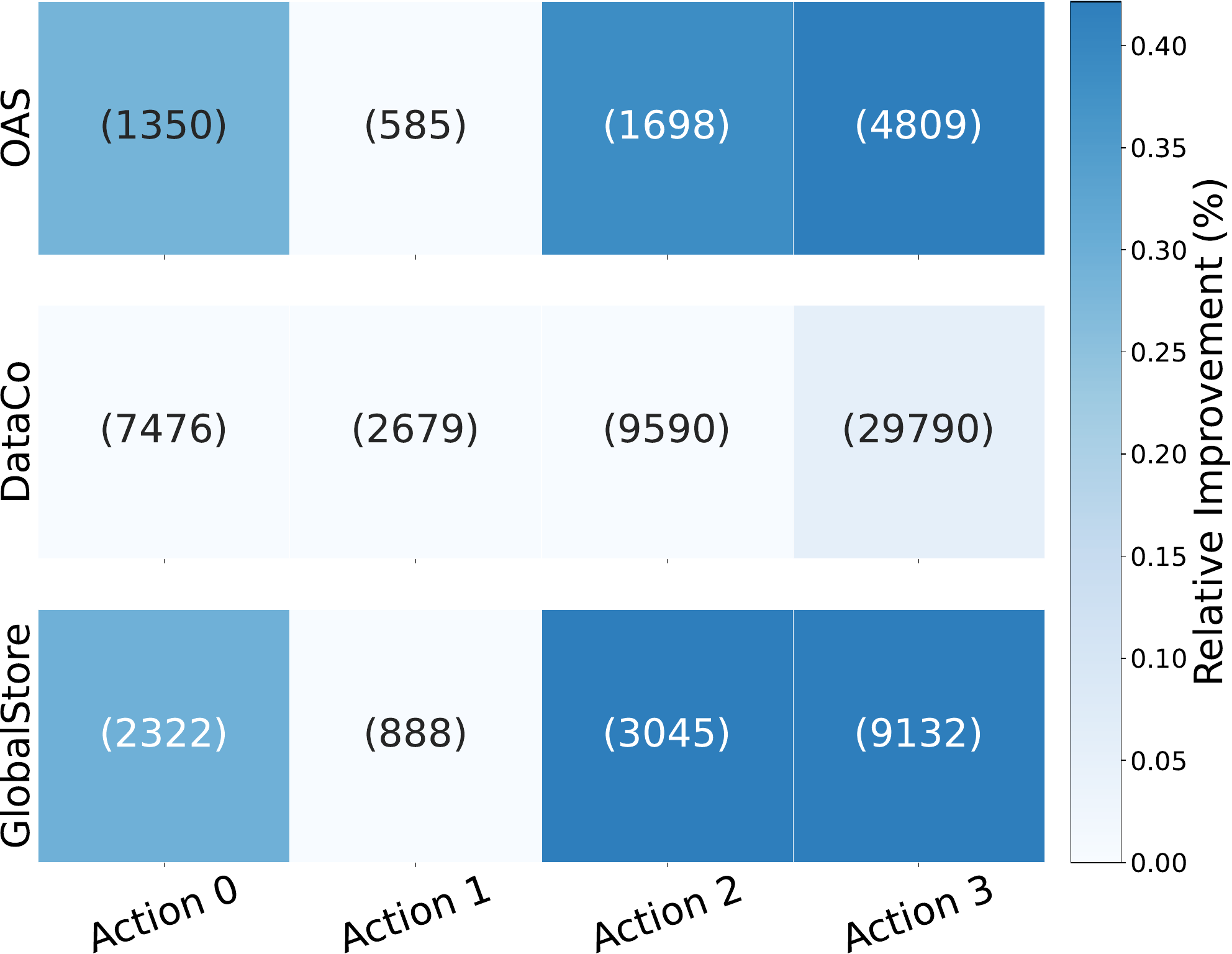}
        \caption{Calibration Heatmap}
        \label{fig:calibration_heatmap}
    \end{subfigure}
    \caption{
    Ablation study of Sim2Act components across three datasets.
    The first three panels report decision robustness under perturbation
    for different module combinations:
    \texttt{None} (S2D), \texttt{+SimCal}, \texttt{+DecPert}, and \texttt{+Both}.
    \texttt{+SimCal} improves robustness in decision-critical regions
    by reducing performance degradation,
    while \texttt{+DecPert} stabilizes decision behavior under uncertainty.
    The calibration heatmap (right) visualizes action-level reliability gains,
    showing that simulator calibration concentrates improvements
    on decision-critical actions.
    }
    \vspace{-0.4cm}
    \label{fig:ablation_heatmap_combined}
\end{figure*}
\noindent\underline{C: Ablation Study of Calibration and Perturbation.}
We assess the individual contributions of the simulator calibration (\texttt{+SimCal})
and decision perturbation (\texttt{+DecPert}),
with neither component enabled (\texttt{None}) as the baseline
(Figure~\ref{fig:ablation_heatmap_combined}).

\noindent\emph{C.1: Ablation Study on Simulator.}
\texttt{+SimCal} meaningfully reduces vulnerability in decision-sensitive regions. For example, on \textbf{DataCo} the profit drop rate decreases from $6.9\%$ to \textbf{4.8\%}, indicating better alignment between simulated outcomes and decision rewards. (See Appendix A
for full numeric tables and per-strength breakdown.)

\noindent\emph{C.2: Ablation Study on Decision-maker }
\texttt{+DecPert} improves robustness under perturbation while keeping nominal performance stable; on \textbf{OAS} the timely-delivery drop is reduced from $0.37\%$ to \textbf{0.22\%}. The combined setup (\texttt{+Both}) yields the most consistent results across datasets. (Full ablation curves and exact numbers are in Appendix A
.)

\paragraph{Q3: Studying Calibration Improvement Heatmap Over Datasets and Actions.}
Figure~\ref{fig:calibration_heatmap} visualizes action-level calibration gains of accuracy after calibration and accuracy before calibration. 
We focus on high-frequency, decision-critical actions: e.g., on \textbf{GlobalStore} the most frequent action (\texttt{Action 3}, 9132 samples) improves from $0.9447$ to \textbf{0.9531}; on \textbf{OAS}, \texttt{Action 3} improves from $0.5694$ to \textbf{0.5718}; on \textbf{DataCo}, \texttt{Action 3} improves from $0.9150$ to \textbf{0.9155}. These representative numbers support the claim that calibration concentrates gains on high-impact actions, more detailed analysis is in Appendix C.

\section{Related Work}

\noindent\textbf{Simulation-based Decision Making and Calibration.}
Simulation is critical for decision-making in high-stakes domains where real-world exploration is prohibitive, such as supply chains and industrial control~\citep{agrawal2025adaptivefewshotlearningafsl, YevgenievichBarykin2020}.
Recently, Sim2Dec~\citep{bai2025supplychainoptimizationgenerative} is a significant step forward by using generative models (LSTM) to approximate real-world environment dynamics and serve as simulator surrogates. 
However, existing environment simulator surrogates suffer from issues caused noisy, drifted, or imperfect data~\citep{Bi_2022}.
Although these methods can  minimize global prediction errors
(e.g., MSE) or maximizing likelihood~\citep{Correia2023, ATANASSOV20081477}, they are limited by the inability to reduce simulation errors in decision-critical regions (i.e., state-action pairs) where  small mispredictions can not flip the entire action ranking orders and impact downstream policy performance~\citep{zhao2021calibratingpredictionsdecisionsnovel, fonteneauBatchModeReinforcement2013}.

\noindent\textbf{Robustness in Model-Based Reinforcement Learning.}
Robustness is essential Model-Based Reinforcement Learning (MBRL), particularly for offline settings where RL agents cannot correct its knowledge via interactions with real environments. Prior studies are two fold: 1) adversarial robustness and 2) conservative regularization. Adversarial methods~\citep{pinto2017robustadversarialreinforcementlearning, zhang2021robustdeepreinforcementlearning} train agents against a worst-case perturbation and are modeled as an opposing player.  Such min-max formulation can lead to overly conservative policies that sacrifice average performance for worst-case safety~\citep{zhang2025stateawareperturbationoptimizationrobust}.
Offline RL methods employ pessimistic regularization~\citep{yang2022rorlrobustofflinereinforcement, LI2025126888} or constrain the policy to stay close to the policy distribution of previous step~\citep{liu2021constrainedmodelbasedreinforcementlearning}. 
These methods however often struggle with distribution shifts and sparse data regions~\citep{NEURIPS2020_0d2b2061}. 
Instead, our method adopts a group-relative perturbation strategy for policy robustness. 

\noindent\textbf{Adversarial Learning and Group-Relative Optimization.}
Our method is inspired by the insights of adversarial learning
and group-relative optimization.
Beyond attack defense, adversarial training can be reformulated to enforce regularization, such as fairness or consistency~\citep{liu2024robustdeepreinforcementlearning, 10889801}.
We adapt this paradigm to simulator calibration, treating the error re-weighting process as an adversarial game.
Besides,  the Group-Relative Proximal Optimization (GRPO)~\citep{shao2024deepseekmathpushinglimitsmathematical, zhang2025grpoleaddifficultyawarereinforcementlearning} has demonstrated remarkable success in large language models and reasoning by evaluating outputs relative to a group mean rather than a fixed baseline, thus, reduces variance and enhances generalization.
We adapt this group-relative concept to perturbation as a tool to robustify policies. 
Moreover, there is no existing work that integrates the adversarial and group-relative concepts to jointly improve simulation surrogates and decision-making policy learning.  
\section{Conclusion Remarks}
Simulation-to-decision learning aims to learn a simulator as environment surrogate and  decision policies by interacting with the surrogate, to avoid testing in real mission-critical systems. 
We study how to robustify both simulator and policy learning. 
Classic methods are limited by the inability to control prediction errors in decision-critical regions that flip entire action rankings; they often are overly regularized to see every uncertainty as threat and discard high-risk high-reward actions. 
We propose the Sim2Act that develops two concepts: action-aligned reweighing-based adversarial calibration and group-relative perturbation, to achieve robustness in both simulator and policy learning. 
Our results find that: 1) correcting high-impact state-action prediction errors is more effective than minimizing average reward loss; 2) group-relative perturbation can robustify policies without downgrading decision reward.
Theoretically, our findings imply that probabilistic consistency is often superior to worst-case adversarial defense.
Practically, our framework enables more reliable digital twins (simulation and decision-maker) deployment in real-world mission-critical domains (e.g., transportation, supply chains) without risking expensive infrastructures.
Our future work is to study physics knowledge guided simulation surrogate and policy learning in complex scientific and engineering systems. 


\bibliographystyle{named}
\bibliography{ijcai26}

\end{document}